%% file: main.tex
\documentclass[10pt,twocolumn,letterpaper]{article}

\PassOptionsToPackage{table}{xcolor}

\usepackage[pagenumbers]{cvpr} 
\usepackage[accsupp]{axessibility}
\usepackage[pagebackref,breaklinks,colorlinks,allcolors=cvprblue]{hyperref}
\usepackage{multirow}

\definecolor{cvprblue}{rgb}{0.21,0.49,0.74}

\def\paperID{1479} 
\def\confName{CVPR}
\def\confYear{2025}

\title{Category-Agnostic Neural Object Rigging}
\newcommand{\methodname}{CANOR\xspace}

\author{
Guangzhao He$^{1,*,\text{†}}$\qquad
Chen Geng$^{1,*}$ \qquad
Shangzhe Wu$^{1,2}$ \qquad
Jiajun Wu$^1$ \\[0.8em]
$^1$Stanford University \qquad
$^2$University of Cambridge
}

\begin{document}
\input{fig/teaser}

{
\let\thefootnote\relax\footnotetext{$^*$Equal contribution. $^{\text{†}}$Work was done when G. He was a visiting student at Stanford University. G. He is currently with Zhejiang University.}
}

\input{sec/0_abstract}    
\input{sec/1_intro}
\input{sec/2_related}
\input{sec/3_method}
\input{sec/4_experiments}
\input{sec/5_conclusions}

\paragraph{Acknowledgments.}
This work is in part supported by ONR YIP N00014-24-1-2117 and NSF RI \#2211258 and \#2338203.

{
    \small
    \bibliographystyle{ieeenat_fullname}
    \bibliography{ref}
}

\end{document}

%% file: fig/teaser.tex
\twocolumn[{
    \maketitle
    \begin{center}
        \captionsetup{type=figure}
        \vspace{-5mm}
        \includegraphics[width=\textwidth]{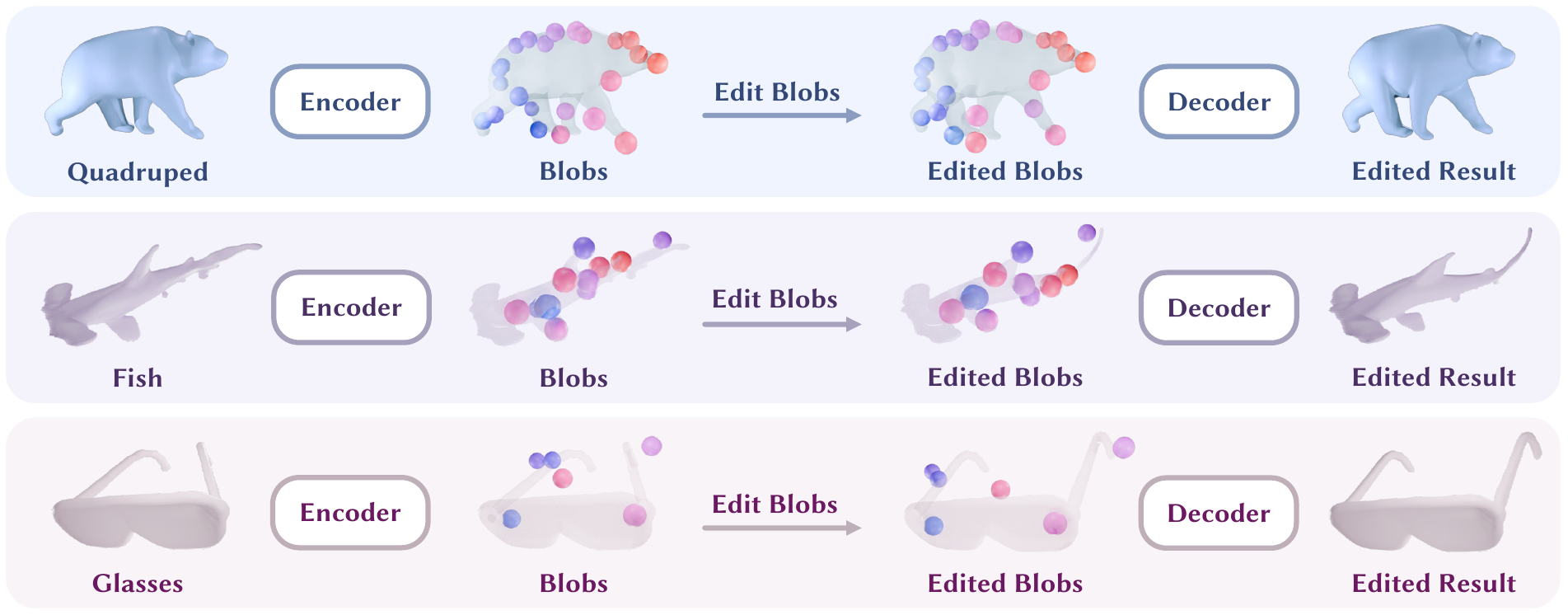}
        \vspace{-5mm}
        \captionof{figure}
        {
            We introduce \textbf{Category-Agnostic Neural Object Rigging (\methodname)}, a novel approach that learns to discover a low-dimensional pose space for dynamic objects. The representation is learned from animated 3D sequences of a deformable object category in an unsupervised fashion without relying on any category-specific expert knowledge. 
            By decomposing each object's geometry into a sparse set of feature-embedded blobs, \methodname enables intuitive manipulation of object poses by editing the blobs. 
            This representation captures interpretable motion structures for a diverse range of dynamic object categories.
        }
        \vspace{5mm}
    \end{center}
}]

%% file: sec/0_abstract.tex
\begin{abstract}

The motion of deformable 4D objects lies in a low-dimensional manifold. To better capture the low dimensionality and enable better controllability, traditional methods have devised several heuristic-based methods, i.e., rigging, for manipulating dynamic objects in an intuitive fashion. However, such representations are not scalable due to the need for expert knowledge of specific categories. Instead, we study the automatic exploration of such low-dimensional structures in a purely data-driven manner. Specifically, we design a novel representation that encodes deformable 4D objects into a sparse set of spatially grounded blobs and an instance-aware feature volume to disentangle the pose and instance information of the 3D shape. With such a representation, we can manipulate the pose of 3D objects intuitively by modifying the parameters of the blobs, while preserving rich instance-specific information. We evaluate the proposed method on a variety of object categories and demonstrate the effectiveness of the proposed framework. Project page: {\url{https://guangzhaohe.com/canor}}.

\end{abstract}

%% file: sec/1_intro.tex
\section{Introduction}

We live in a dynamic 4D world populated by diverse, ever-moving beings --- not just humans, but also pets, wild animals, and other dynamic entities that can move and deform.
Modeling and understanding the \textit{structure} of motion across different categories of deformable objects has been a long-standing challenge in Computer Graphics and 3D Computer Vision, with applications in character animation, 4D reconstruction, and AR/VR.

One fundamental property of the motion structure shared by almost all dynamic objects is their inherent low-dimensionality, often captured using various \textit{rigging} representations~\cite{89rigging}. 
Historically, extensive efforts have been dedicated to crafting such representations with domain-specific expertise. For commonly-studied categories, such as humans, domain-specific skeleton structures and skinning methods \cite{BogoKLG0B16, kavan2007skinning} have been developed. These structured representations significantly facilitate downstream tasks by offering an interpretable motion structure and reliable correspondences across different dynamic states.

While domain-specific representations have been successful for certain dynamic object categories, their development requires extensive expertise, making it impractical to design such structures for every categories of interest. Recently, several methods have attempted to discover similar structures for other dynamic categories without extensive manual intervention \cite{banmo,lassie,magicpony}. However, most of these approaches still rely on some level of category-specific prior knowledge, which limits their applicability to generic categories.

In this paper, we study the automatic exploration of rigging representations for any dynamic object category, using minimal 3D data, and without any category-specific prior knowledge. Given several animated 3D shape sequences of instances from a certain deformable object category, such as \textit{bears}, 
we propose an algorithm that extracts the shared pose space within the category. This exploration process is entirely category-agnostic, assuming no prior knowledge of the category, correspondences, or other instance-specific information.

To achieve this, we exploit the key property of rigging representations: their low dimensionality. Drawing inspiration from traditional skeleton-based motion modeling, we introduce a sparse set of spatial-grounded blobs to represent the dynamic poses of moving instances. Given the 3D shape of an instance in a specific pose, we train an encoder to decompose it into dynamic blobs encoding pose information along with instance-aware features capturing instance-specific details. These disentangled bottleneck representations can be further decoded back into the 3D shape.

Once this representation is obtained, we can manipulate the deformable objects in an intuitive manner, where the user can directly drag the extracted blobs to modify object's pose. Moreover, the learned representation reveals a low-dimensional and intuitive structure underlying the high-dimensional deformation space for dynamic objects. To demonstrate these, we apply our method to several diverse object categories that are rarely addressed by prior work, and show significant improvements over state-of-the-art baselines.

In summary, our contributions are:

\begin{itemize}
\item We explore a novel task of exploring a category-specific rigging representation in a category-agnostic manner.
\item We develop a representation that automatically encodes the 3D shape information into spatially grounded blobs and instance-aware features.
\item We evaluate the proposed pipeline on several different deformable object categories and demonstrate significant improvements compared to the State of the Art.
\end{itemize}

%% file: sec/2_related.tex
\input{fig/pipeline}
\section{Related work}

\paragraph{Traditional Rigging Representations.}
Crafting expressive yet intuitive rigging representations for deformable objects remains a fundamental challenge in the character animation community. For rigging humanoid characters or mammals, the most intuitive method is to annotate their skeletal structures and associated skinning weights~\cite{89rigging,loper2023smpl,smal}. However, this process typically demands substantial manual efforts from artists. Recent works explored on automating the annotation process ~\cite{rignet,xu2019rig,morig,baran2007automatic}; however, these models often exhibit limited generalization capacities due to insufficient training data. For object categories lacking hierarchical skeleton structures, such as faces, researchers have explored machine learning approaches to develop low-dimensional parametric representations from large databases of aligned shapes~\cite{neuralfacerigging,3dmm,bailey2020fast,3dmmsurvey}. Nevertheless, these approaches typically require substantial amount of high-quality training data, making them challenging to scale for the generic categories considered in this work. 

\paragraph{Neural Rigging Representations.} Beyond rigging representations with explicit analytical decoding processes, recent research has extensively explored neural-based rigging representations. These approaches leverage deep neural networks to decode latent pose representations into detailed shapes. Our work falls into this category. Within this paradigm, keypoints or handles have emerged as a common control modality~\cite{keypointdeformer,mehmet2015,deepmetahandles,yoo2024neural,liu2023reart}, which are conceptually similar to the blobs utilized in this work. However, prior works have primarily focused on predicting the deformation of static shapes, whereas our method directly generates posed shapes and targets dynamic objects. Neural Deformation Graphs~\cite{BozicPZTDN21} optimizes node-based representation for rigging dynamic objects, but require a sequence of 3D SDFs as input and lack a learned categorical prior. In contrast, our method performs amortized inference to predict rigging representation directly from a static 3D shape. Another line of work represent complex shapes using learned latent codes without explicit spatial locations ~\cite{neuralcages,wang20193dn,palafox2021npms,zhou2020unsupervised,palafox2021spams,giebenhain2023nphm}. While effective, these representations typically lack interpretability, while our representation encodes the spatial layout of the posed shape in a more intuitive manner.

\paragraph{4D Representations.} Prior work on 4D representations often employ low-dimensional structures to regulate motion, including part-based~\cite{yao2022lassie,yang2022banmo,yao2023hi}, skeleton-based~\cite{magicpony,nb,instantnvr,lei2024gart,wu2023dove}, phase-based~\cite{phasepgf}, and node-based~\cite{BozicPZTDN21}. While most of these works rely on rule-based scheme to decode latent into posed shapes, our approach learns a neural decoder directly from data. Other 4D representations~\cite{wang2024shape,lei2024mosca,luiten2024dynamic,stearns2024dynamic} directly model high-dimensional flow between different posed shapes. However, these methods often require dense inputs~\cite{luiten2024dynamic} or rely on manually defined regularizations~\cite{arap,yoo2024plausible,huang2021arapreg}.

\paragraph{Mid-level Neural Representations.} Our work is also inspired by the large body of research that leverages mid-level neural representations to model visual contents. Most existing approaches focus on capturing scene-level object layouts~\cite{epstein2022blobgan,carson1999blobworld}. Similar to our method, BlobGAN~\cite{epstein2022blobgan} employs blobs as a mid-level neural representation; however, their work is limited to 2D images of indoor scenes. Deep Latent Particles~\cite{deeplatentparticles} also operates in the 2D domain, demonstrating applications in manipulating human faces using these representations.

%% file: fig/pipeline.tex
\begin{figure*}[t]
    \centering
    \includegraphics[width=1.0\textwidth]{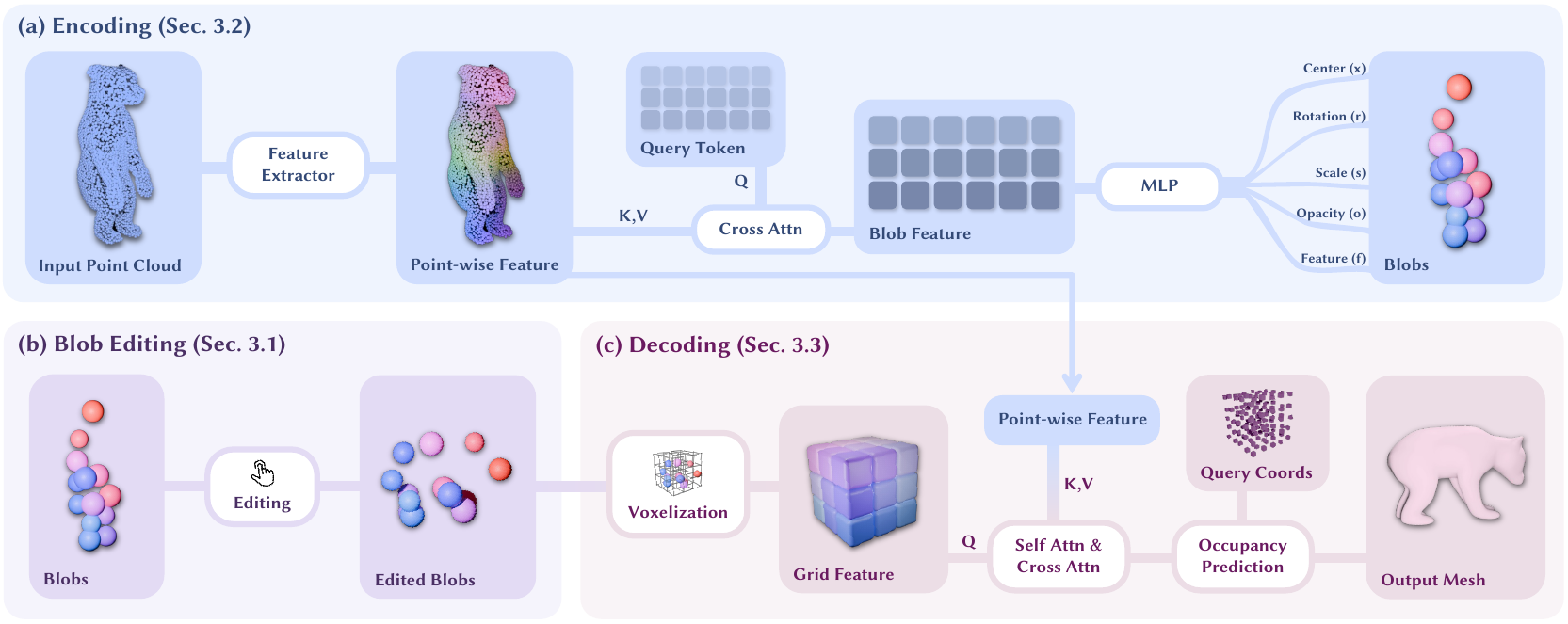}
    \vspace{-17pt}
    \caption{
        \textbf{Overview of our proposed pipeline.} We use a set of feature-embeded blobs to represent the pose space of deformable objects (\cref{method:symbols}). 
        The encoder takes a point cloud as input and maps it into blobs using a learnable codebook of query tokens that cross-attend with semantic point-wise features~(\cref{method:encoding}).
        Once generated, these blobs can be edited by users to adjust the object's pose. The edited blobs are then voxelized into a feature volume and decoded back to a 3D volume using a transformer architecture~(\cref{method:decoding}).
        Finally, the system query the decoded volume with sampled 3D coordinates to predict occupancy values, which are used to extract the edited mesh.
    }
    \label{figure:pipeline}
    \vspace{-12pt}
\end{figure*}

%% file: sec/3_method.tex
\section{Method}
\label{method}

Given an instance shape from a deformable object category in the form of a point cloud $\mathbf{P}\in\mathbb{R}^{{n_p}\times3}$, such as \textit{bears}, \textit{fish}, or \textit{laptops}, our goal is to predict a structured and interpretable representation $\mathcal{B}$ that captures the dynamic \textit{poses} of the object. 
This representation can be intuitively edited by users to animate or re-pose the 3D object. 
Inspired by the concept of skeleton-based rigging representations~\cite{89rigging,loper2023smpl}, we introduce a sparse set of \textit{blobs} to implicitly encode the spatial structure of dynamic objects. Each blob $\mathbf{b}_i\in\mathcal{B}$ is an anisotropic sphere parameterized by its position, rotation, radius, and feature, which collectively indicate how a semantic part of a dynamic object is positioned at a certain pose.

We learn to discover such a representation in an unsupervised manner. Given the input point cloud $\mathbf{P}$, we define an encoder $\mathcal{E}(\mathbf{P})=\mathcal{B}$ that maps $\mathbf{P}$ to a sparse set of blobs representing its current pose. 
A decoder $\mathcal{D}(\mathcal{B};\mathbf{P})$ is trained to reconstruct the object shape as a mesh. 
The position and rotation of each blob can be edited to create a novel dynamic pose $\mathcal{B}'$. This modified pose $\mathcal{B}'$ can be subsequently decoded using $\mathcal{D}(\mathcal{B}';\mathbf{P})$ to generate the re-posed shape of $\mathbf{P}$, as illustrated in~\cref{figure:pipeline}.

The following subsections provide a detailed description of our proposed model. 
We begin by introducing the design of the blob-based representation (\cref{method:symbols}).
In~\cref{method:encoding} and~\cref{method:decoding}, we detail the architectures for $\mathcal{E}$ and $\mathcal{D}$, respectively.
Finally, the training details are discussed in~\cref{method:optimization}.

\subsection{Representing Object Pose as Blobs}
\label{method:symbols}

In this subsection, we describe our rigging representation in the form of a set of blobs. 

\paragraph{Blob Parametrization.} Each blob in the set captured both spatial and local semantic information corresponding to a specific part of the dynamic object. A blob $\mathbf{b}$ is defined as a feature-embedded anisotropic sphere:

\begin{equation}
    \mathbf{b}=(\mathbf{x},\mathbf{r},\mathbf{s},\mathbf{o},\mathbf{f}),
\end{equation}
where $\mathbf{x}\in\mathbb{R}^3$ denotes the center position of the blob, $\mathbf{r}\in\mathbb{H}$ represents its orientation as a rotation quaternion, $\mathbf{s}\in\mathbb{R}$ is the radius of the sphere, $\mathbf{o}\in[0,1]$ indicates the blob's opacity indicating the activation level, and $\mathbf{f}\in\mathbb{R}^d$ is a feature vector that encodes local semantic information used for shape decoding.
With a set of such blobs $\mathcal{B}=\{\mathbf{b}_i\,|\,i=1,...,n_b\}$, we can decompose the representation into pose-dependent parameters $\mathcal{B}_P=\{(\mathbf{x}_i,\mathbf{r}_i)\,|\,i=1,...,n_b\}$ and identity-dependent parameters $\mathcal{B}_I=\{(\mathbf{s}_i,\mathbf{o}_i,\mathbf{f}_i)\,|\,i=1,...,n_b\}$.

\paragraph{Remarks.}
Blobs offer a flexible, category-agnostic alternative to traditional skeleton-based representations~\cite{89rigging} for modeling object poses. Unlike skeletons, which impose rigid hierarchical structures and often require manual design and tuning for different categories, blobs model objects as collections of semi-rigid parts. This makes them easier to learn and generalize across diverse shapes and motion patterns.
Moreover, the blob-based design enables intuitive and flexible pose editing. For instance, users can manipulate blob positions ($\mathbf{x}$) and orientations ($\mathbf{r}$) to adjust the object's pose, or modify radius ($\mathbf{s}$) to resize specific parts.

\subsection{From Shape to Blobs}
\label{method:encoding}

Given an input point cloud $\mathbf{P}$, we define a feed-forward encoding process $\mathcal{E}$ that maps $\mathbf{P}$ into a set of blobs $\mathcal{B}$ as described above. 

$\mathcal{E}$ consists of two components: $\mathcal{E}_P(\mathbf{P})=\mathcal{B}_P$, which predicts pose-related parameters, and $\mathcal{E}_I(\mathbf{P})=\mathcal{B}_I$, which predicts identity-related parameters. This can be formalized as:
\begin{equation}
    \mathcal{E}(\mathbf{P})=
    \{(\mathcal{E}_P(\mathbf{P})[i],\mathcal{E}_I(\mathbf{P})[i])\,|\,i=1,...,n_b\}.
\end{equation}

Both encoding processes $\mathcal{E}_P$ and $\mathcal{E}_I$ begin with a shared feature extractor that computes point-wise features $\mathbf{F}\in\mathbb{R}^{{n_p}\times{d}}$ from the input point cloud $\mathbf{P}$.  

Next, we perform cross-attention between the point-wise features $\mathbf{F}$ and a learnable codebook $\mathcal{Q}=\{\mathbf{q}_i\,|\,i=1,...,n_b\}$, where each code corresponds to a distinct blob. This yields attention weights $\mathbf{W}\in\mathbb{R}^{{n_b}\times{n_p}}$ between the codebook tokens and input points. The attention weights are used to compute two sets of aggregated feature vectors $\mathcal{F}_P=\{\mathbf{f}_P^i\,|\,i=1,...,n_b\}$ for pose and $\mathcal{F}_I=\{\mathbf{f}_I^i\,|\,i=1,...,n_b\}$ for identity.

Finally, the blob parameters $\mathcal{B}_P$ and $\mathcal{B}_I$ are regressed from $\mathcal{F}_P$ and $\mathcal{F}_I$, respectively.

Further details of each component are discussed below.

\paragraph{Feature Extractor.} 
To distinguish different semantic parts of the object, maintain pose consistency, and enable accurate shape reconstruction during decoding, the point-wise features $\mathbf{F}$ must encode both semantic and geometric information.

We adopt PointTransformer~\cite{ZhaoJJTK21} as our feature extractor due to its strong performance in capturing consistent, expressive, and discriminative features. This capability allows it to recover high-quality details and effectively handle challenging symmetric structures.

\paragraph{Shape Encoding with a Learned Codebook.}

We aggregate the above-mentioned point-wise feature into a low-dimensional sparse set by learning a shared codebook $\mathcal{Q}$ and performing cross-attention-based feature aggregation. Each token in the codebook corresponds to a distinct blob that captures a specific semantic part. The codebook is learned jointly with other components and shared across object identities to ensure consistency.

To compute the cross-attention-based feature aggregation, we calculate the attention map $\mathbf{W}$ between each token $\mathbf{q}\in\mathcal{Q}$ and each point in the input.
The attention weight $\mathbf{W}[i,j]$ between the $i$-th code and $j$-th point is given by:
\begin{equation}
    \mathbf{W}[i,j]=\frac{\exp{(\mathcal{Q}[i]\cdot{\mathbf{F}[j]]}^T/\sqrt{d})}}{\sum\nolimits_{j=1}^{n_p}\exp{(\mathcal{Q}[i]\cdot{\mathbf{F}[j]}^T/\sqrt{d})}}.
\end{equation}

The attention weights $\mathbf{W}$ are used to compute two different sets of feature vectors: $\mathcal{F}_P$ for pose and $\mathcal{F}_I$ for identity.

For the pose-related features $\mathcal{F}_P$, we aggregate the positional encodings $\gamma(\cdot)$~\cite{MildenhallSTBRN20} of the point coordinates to capture spatial relationships between blobs:
\begin{equation}
    \mathcal{F}_P[i]=\sum\nolimits_{j=1}^{n_p}\mathbf{W}[i,j]\cdot\gamma(\mathbf{P}[j]).
\end{equation}

To extract identity features $\mathcal{F}_I$, which require detailed geometric information, we concatenate the extracted feature $\mathbf{F}$ and point-wise positional encoding $\gamma(\mathbf{P})$, then pass the fused vector through an MLP $\phi$:
\begin{equation}
        \mathcal{F}_I[i]=\sum\nolimits_{j=1}^{n_p}\mathbf{W}[i,j]\cdot \phi(\mathbf{F}[j]\oplus\gamma(\mathbf{P})[j]),
\end{equation}
where $\oplus$ denotes the concatenation operator.

Finally, the aggregated feature vector sets $\mathcal{F}_P$ and $\mathcal{F}_P$ are then passed through a group of five shallow MLPs, each responsible for regressing a specific type of blob parameter.

\subsection{From Blobs Back to Shape}
\label{method:decoding}

After encoding the input shape into a set of blobs $\mathcal{B}$, these blobs can be explicitly edited to obtain $\mathcal{B}'$, representing potential pose changes.
To reconstruct the re-posed object corresponding to $\mathcal{B}'$, we first voxelize the blobs into a feature volume $\mathbf{V}\in\mathbb{R}^{({h}\times{w}\times{l})\times{d}}$, where $h$, $w$ and $l$ are the dimensions of the volume.
The feature volume is then iteratively refined through a series of self-attention layers~\cite{VaswaniSPUJGKP17}, and subsequently decoded into an occupancy field that represents the final 3D shape.
During this iterative refinement, the volume is also conditioned on the extracted features $\mathbf{F}$ from the encoder to better preserve object identity. We describe the details below.

\paragraph{Blob Feature Volume.}
Instead of directly passing blob parameters to the decoder, we employ a differentiable voxelization process to ensure that each blob's parameters $\mathbf{x}$, $\mathbf{r}$, $\mathbf{s}$ and $\mathbf{o}$ have explicit and interpretable effects.

To voxelize the blobs, we first construct a 3D grid of coordinates $\mathbf{G}\in\mathbb{R}^{({h}\times{w}\times{l})\times3}$. The feature at each grid point $\mathbf{g}_i$ is computed as a weighted summation of blob features $\mathbf{f}_j$ w.r.t.\ weight distribution $w_{ij}$: 

\begin{equation}
    \mathbf{F}_G[i]=\frac{\sum\nolimits_{j=0}^{n_b} w_{ij}\cdot\mathbf{{f}}_{j}}{\sum\nolimits_{j=0}^{n_b}
    w_{ij}+\epsilon},
\end{equation}
where $\epsilon$ is a small constant to prevent numerical instability. The weights $w_{ij}$ capture the influence of blob $j$ on grid point $i$, and are defined as:

\begin{equation}
    w_{ij}=\mathbf{o}_j \cdot \text{exp}{(-c\cdot (\frac{\mathbf{g}_i-\mathbf{x}_j}{\mathbf{s}_j})(\frac{\mathbf{g}_{i}-\mathbf{x}_j}{\mathbf{s}_j})^T)},
\end{equation}
where $c$ is a constant that controls the softness of the kernel; we set $c=1$ in all our experiments. 

As a special case, $w_{i0}$ and $f_{i0}$ represents a learnable background weight and feature, respectively, which are shared across all inputs.

\paragraph{Augmenting Blob Features.}

\label{method:augmentation}

In practice, using only blob features to construct the feature volume may lead to a loss of fine-grained details in reconstruction due to the compactness of the blob representation. To address this, we propose to augment blob features prior to voxelization, enhancing their expressiveness.

Specifically, we enrich each blob feature $\mathbf{f}_j$ with point-specific embeddings $\mathcal{F}$. The augmented feature $\mathbf{\tilde{f}_{ij}}\in\mathbb{R}^d$ for the $i$-th grid $\mathbf{g}_i$ and blob $\mathbf{b}_j$ is computed as:
\begin{equation}
    \mathbf{\tilde{f}}_{ij}=\mathbf{f}_j+\mathbf{W}_\theta(\gamma(\frac{\mathbf{g}_i-\mathbf{x}_j}{\mathbf{s}_j}\cdot \mathrm{R}(\mathbf{r}_j))),
\end{equation}

where $\mathbf{W}_\theta$ denotes an MLP with parameters $\theta$ and $\mathrm{R}(\cdot)$ denotes the conversion from a quaternion to a $3\times{3}$ rotation matrix.

Substituting $\mathbf{f}_{ij}$ with $\tilde{\mathbf{f}}_{ij}$ in the voxelization process described above, we obtain the augmented feature volume $\tilde{\mathbf{F}}_G$:

\begin{equation}
    \mathbf{\tilde{F}}_G[i]=\frac{\sum\nolimits_{j=0}^{n_b} w_{ij}\cdot\mathbf{\tilde{f}}_{ij}}{\sum\nolimits_{j=0}^{n_b}
    w_{ij}+\epsilon}.
\end{equation}

\paragraph{Feature Decoding.}
Given the augmented grid features $\mathbf{\tilde{F}}_G$, we follow recent shape auto-encoding methods~\cite{ZhaoLCZWCFCYG23, ZhangTNW23} to iteratively process them through a series of self-attention layers to produce $\mathbf{F}_G'$.
We add positional encodings $\gamma(\mathbf{G})$ to the features and treat each as an individual token within the attention mechanism.

To better condition the predicted shape on the identity of the input, we further incorporate cross-attention layers between the grid features and the point-wise features $\mathbf{F}$ extracted during encoding. This conditioning was found to significantly boost the preservation of fine-grained details, as demonstrated in the ablation studies. Addtional details on the decoding architecture are provided in the supplementary material.

\paragraph{Occupancy Prediction.}
Following prior works~\cite{ZhaoLCZWCFCYG23, ZhangTNW23}, for any queried coordinate $\mathbf{x}_q\in\mathbb{R}^3$, our network outputs an occupancy value $o\in[0,1]$ using the decoded grid features $\mathbf{F}'_G$. We first perform cross-attention between the postional encoding of $\mathbf{x}_q$ and $\mathbf{F}'_G$, and then pass it through an MLP to obtain the final occupancy. We use Marching Cubes~\cite{LorensenC87} to extract the final mesh.

\subsection{Training}
\label{method:optimization}
\input{fig/training}

\paragraph{Training Strategy.}
The pipeline described above takes an object point cloud $\mathbf{P}$ and user-edited blobs $\mathcal{B}'$ as input and predicts the re-posed mesh $\mathcal{M}'$ of the object.
However, directly obtaining the data tuple $(\mathbf{P},\mathcal{B}',\mathcal{M}')$ is impractical.
Instead, we sample two distinct frames of a deformation sequences and obtain their meshes $\mathcal{M}$ and $\mathcal{M}'$ as inputs, ensuring that they share the same object identity but differ in pose.
To derive the edited blobs $\mathcal{B}'$, 
we separately predict pose-related parameters $\mathcal{B}_P$ from $\mathcal{M}'$ and identity-related parameters $\mathcal{B}_I$ from $\mathcal{M}$. Combining these parameters yields blobs that represent the identity of $\mathcal{M}$ with the pose of $\mathcal{M}'$, thus capturing the pose change from $\mathcal{M}$ to $\mathcal{M}'$.

Formally, the training tuple is expressed as:
\begin{equation}
    (\mathbf{P}_\mathcal{M},\{(\mathcal{E}_I(\mathbf{P}_{\mathcal{M}})[i],\mathcal{E}_P(\mathbf{P}_{\mathcal{M}'})[i])\},\mathcal{M}'),
\end{equation}
where $\mathbf{P}_\mathcal{M}$ and $\mathbf{P}_\mathcal{M}'$ denote the sampled point cloud from mesh $\mathcal{M}$ and $\mathcal{M}'$, respectively.
The difference between the training and inference input is illustrated in~\cref{figure:training}.

\input{tab/comparison}
\input{fig/comparison}

\paragraph{Training Objectives.}

We use binary cross entropy loss $\mathcal{L}_\text{recon}$ to supervise the proposed model in an end-to-end manner. During training, we apply a sampling mechanism that biases towards near-surface points with a ratio of $\alpha_\text{ns}$ to enhance high-frequency surface details of the reconstruction. 
To ensure robust convergence, we additionally regularize the summed grid weights $\mathbf{W}_G = \{ \Sigma_{j=1}^{n_b} {w_{ij}}\,|\, i=1,...,h\times w\times l\}$ during voxelization.
The regularization term $\mathcal{L}_\text{vox}$ is defined as the cosine similarity between $\mathbf{W}_G$ and the GT occupancy of the grid $\mathbf{O}_G$.

The complete training objective is defined as:
\begin{equation}
    \mathcal{L}=\mathcal{L}_\text{recon}+\lambda_\text{vox}\mathcal{L}_\text{vox},
\end{equation}
where $\lambda_\text{vox}$ is a hyper-parameter. 
We refer the readers to the supplementary material for more details on training.

%% file: fig/training.tex
\begin{figure}[t]
    \centering
    \vspace{8pt}
    \includegraphics[width=0.45\textwidth]{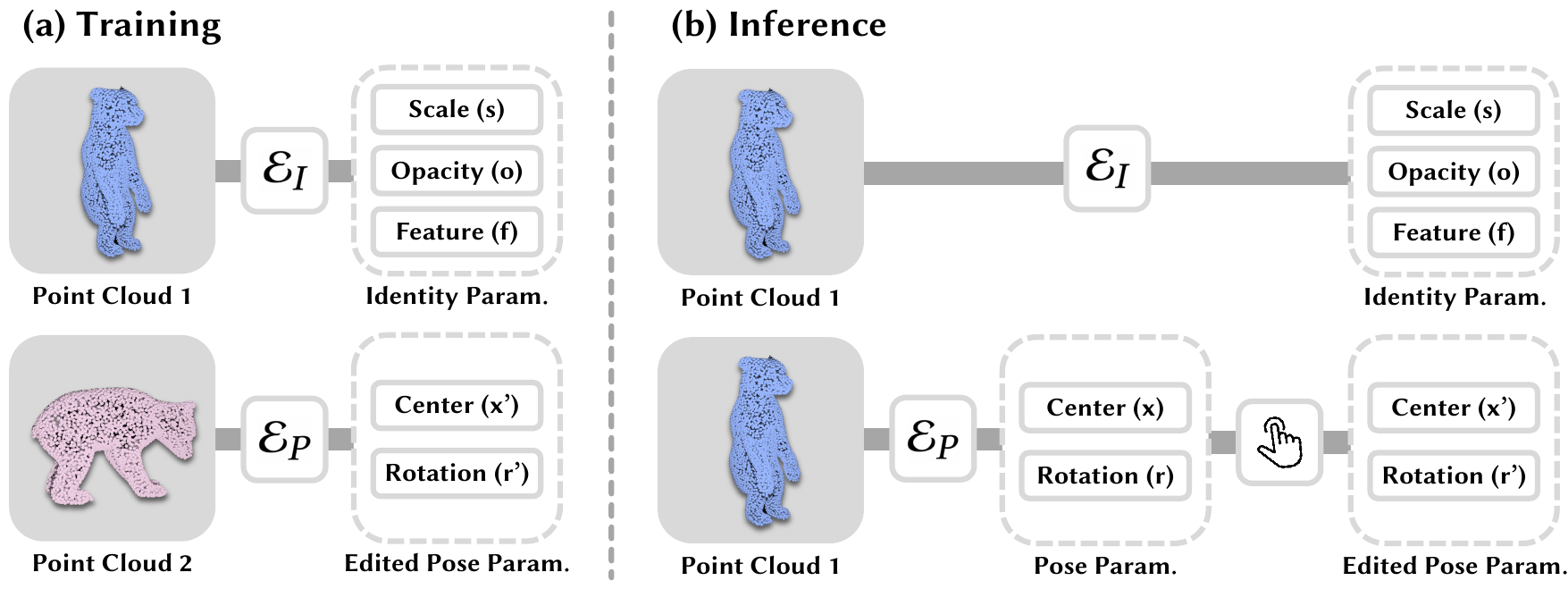}
    \vspace{-5pt}
    \caption{
        \textbf{Difference in training and inference inputs.}
        During training, we sample two point clouds of the same identity but with different poses to separately predict the identity-related blob parameters $\mathcal{B}_I$ and pose-related parameters $\mathcal{B}_P$.
        This setup enables $\mathcal{B}_P$ to simulate an edited pose resulting from user edits.
        During inference, both $\mathcal{B}_I$ and $\mathcal{B}_P$ are predicted from a single point cloud.
        The user can then explicitly edits $\mathcal{B}_P$ to represent the desired pose change.
    }
    \label{figure:training}
    \vspace{-6pt}
\end{figure}

%% file: tab/comparison.tex
\definecolor{colorfirst}{rgb}{.866,.945, 0.831}
\definecolor{colorsecond}{rgb}{1, 0.98, 0.83}
\definecolor{colorthird}{rgb}{0.76, 0.87, 0.92}

\newcommand{\cellfirst}{\cellcolor{colorfirst}}
\newcommand{\cellsecond}{\cellcolor{colorsecond}}
\newcommand{\cellthird}{\cellcolor{colorthird}}

\newcommand{\textfirst}{\colorbox{colorfirst}}
\newcommand{\secondtext}{\colorbox{colorsecond}}
\newcommand{\thirdtext}{\colorbox{colorthird}}

\begin{table*}[t]
    \centering
    \caption{%
        \textbf{Quantitative comparison on DeformingThings4D~\cite{LiTTZN21}, FaMoS~\cite{BolkartLB23}, Fish~\cite{sketchfab}, and \textit{Refrigerator} and \textit{Eyeglasses} from articulated objects dataset Shape2Motion~\cite{WangZSCZ019}.}
        Metrics are averaged over all sequences. 
        IoU, Chamfer Distance $L_1$ and $L_2$ are reported. 
        Our method demonstrates significant improvement compared to the state of the arts. Green and yellow cell colors indicate the best and the second best results, respectively.
    }
    \label{tab:comparison}
    \vspace{-7pt}
    \centering\small
    \setlength{\tabcolsep}{6.5pt}
    \resizebox{\textwidth}{!}{
    \begin{tabular}{lccccccccccccccc} %
        \toprule
                                                   \multirow{2}{*}{Method}
                                                   & \multicolumn{3}{c}{DeformingThings4D~\cite{LiTTZN21}} & \multicolumn{3}{c}{FaMoS~\cite{BolkartLB23}} & \multicolumn{3}{c}{Fish~\cite{sketchfab}}  & \multicolumn{3}{c}{Refrigerator} &
                                                   \multicolumn{3}{c}{Eyeglasses}\\
        \cmidrule(lr){2-4} \cmidrule(lr){5-7} \cmidrule(lr){8-10} \cmidrule(lr){11-13} \cmidrule(lr){14-16}
                                                   & IoU $\uparrow$ & $CD_{1}$ $\downarrow$ & $CD_{2}$ $\downarrow$  & IoU $\uparrow$ & $CD_{1}$ $\downarrow$ & $CD_{2}$ $\downarrow$  & IoU $\uparrow$ & $CD_{1}$ $\downarrow$ & $CD_{2}$ $\downarrow$ & IoU $\uparrow$ & $CD_{1}$ $\downarrow$ & $CD_{2}$ $\downarrow$ & IoU $\uparrow$ & $CD_{1}$ $\downarrow$ & $CD_{2}$ $\downarrow$ \\
        \midrule
        KeypointDeformer \cite{Jakab0M0SK21} & 0.536 & 0.060 & 0.044 & \cellsecond 0.923 & 0.029 & \cellsecond 0.020  & 0.499 & 0.062 & 0.047 & 0.744 & 0.060 & 0.042 & 0.452 & 0.047 & 0.034 \\
        NeuralDeformationGraph \cite{BozicPZTDN21} & \cellsecond 0.875 & \cellsecond 0.020 & \cellsecond 0.013 & 0.800 & \cellsecond 0.019 & \cellfirst 0.013 & 0.686 & \cellsecond 0.040 & \cellsecond 0.030 & \cellsecond 0.869 & \cellsecond 0.046 & \cellsecond 0.034 & \cellfirst 0.791 & \cellsecond 0.024 & \cellsecond 0.016 \\
        SkeRig \cite{Deng14} & 0.802 & 0.057 & 0.041 & 0.790 & 0.045 & 0.031 & \cellsecond 0.782 & 0.049 & 0.035 & 0.803 & 0.105 & 0.074 & 0.544 & 0.128 & 0.096 \\
        \midrule
        Ours & \cellfirst 0.937 & \cellfirst 0.017 & \cellfirst 0.011 & \cellfirst 0.960 & \cellfirst 0.018 & \cellfirst 0.013 & \cellfirst 0.860 & \cellfirst 0.024 & \cellfirst 0.017 & \cellfirst 0.903 & \cellfirst 0.031 & \cellfirst 0.022 & \cellsecond 0.770 & \cellfirst 0.020 & \cellfirst 0.014 \\
        
        \bottomrule
    \end{tabular}
    }
    \vspace{-10pt}
\end{table*}

%% file: fig/comparison.tex
\begin{figure*}[t]
    \centering
    \includegraphics[width=1.0\textwidth]{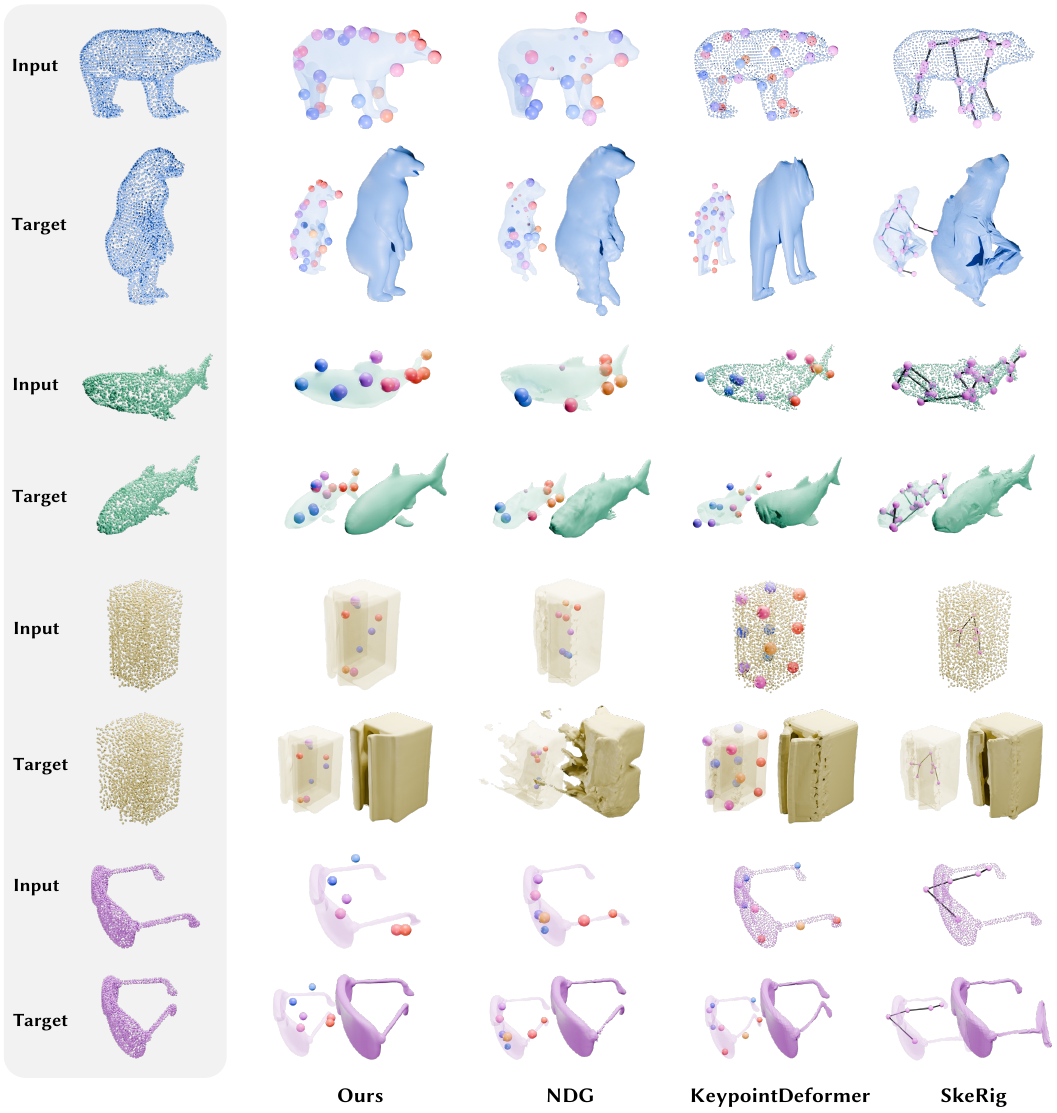}
    \vspace{-16pt}
    \caption{
        \textbf{Qualitative results.} We show qualitative results for different rigging representations across four object categories. Our approach outperforms state-of-the-art methods on both modeling object motion and generating high-quality surface meshes.
    }
    \label{figure:comparison}
    \vspace{-5pt}
\end{figure*}

%% file: sec/4_experiments.tex
\section{Experiments}
\subsection{Implementation Details}
We implement the proposed framework in PyTorch and train it end-to-end using AdamW ~\cite{LoshchilovH19} with a learning rate of $5\mathrm{e}{-4}$. We use 1-Cycle scheduler with linear annealing to accelerate training.
We set near-surface sampling ratio $\alpha_\text{ns}$ to 0.0 for the first 200k iterations to ensure stable gradients for blob initialization, gradually increasing it to 0.5 between 200k and 250k iterations to capture high-frequency details,  and finally set it to 0.8 to accelerate convergence.
Providing good initializations of blobs before increasing near-surface sampling ratio proved curcial for achieving better spatial distributions of blobs and avoiding local minima.
Our training typically converges after 300k iterations, requiring approximately 7 days on 2 RTX A6000 GPUs.

We set the number of blobs
\footnote{We only need to provide an upper bound of the number of blobs. } 
$n_b$ to range from 8 to 24 depending on the exact category being modeled. During voxelization, we use a spatial resolution of $8\times 8 \times 8$, which we find sufficient to capture rich identity details. All attention modules contain 8 self-attention layers implemented  using memory-efficient attention~\cite{abs-2112-05682}.

\subsection{Evaluating Learned Rigging Representation}

\paragraph{Experiment Setup.}
To evaluate the effectiveness of our learned rigging representation and compare it with the state-of-the-art methods, we assess each method's capability to re-pose a source shape to match a target shape.
We then calculate similarity metrics between the re-posed and target shapes.
For our method, we directly regress blob positions and orientations from the target shape, using them as pose-related parameters for re-posing. For baselines that do not support feed-forward pose regression, we fix their learned rigging parameters and optimize only pose-related parameters to evaluate their maximum achievable accuracy.

\paragraph{Datasets.}
We evaluate the proposed method on a diverse set of dynamic object categories to demonstrate its ability to learn rigging representations without relying on category-specific priors. The data includes a quadruped animal dataset DeformingThings4D~\cite{LiTTZN21}, a human facial expression dataset FaMoS~\cite{BolkartLB23}, two articulated object categories \textit{Refrigerator} and \textit{Eyeglasses} from Shape2Motion~\cite{WangZSCZ019}, and a custom \textit{Fish} dataset gathered and curated from Sketchfab~\cite{sketchfab}. 
Each dataset contains animated sequences of dynamic objects within a single category. 
For each dataset, we leave out the longest 5 sequences for testing, since some of the baselines methods require hours of optimization for each sequence.
We then split the rest of the datasets into training and validation sets to train amortized-inference methods including ours.
During testing, we use the first frame from the tested sequence as the source shape and evaluate performance based on the similarity between the re-posed source shape and target shapes from subsequent frames.

\paragraph{Baselines.}
We compare our method against 3 state-of-the-art baselines with distinct rigging representations. \textbf{SkeRig}~\cite{Deng14} is a skeleton-based method that optimizes skeleton structures and skinning weights using a small number of frames from a deformable object. It requires additional dense correspondence across frames as input, which may not be available in our training data. Therefore, we estimate per-frame correspondences using a non-rigid registration pipeline NDP~\cite{0143H22} and provide them as part of the input to this baseline.
To extract the skeleton, we uniformly sample 10 frames from the training set. Once the skeleton is obtained, we freeze its structure and optimize only the bone transformations using differentiable forward kinematics to best fit the target shapes. 
\textbf{KeypointDeformer}~\cite{Jakab0M0SK21} is a keypoint-based method that predicts a set of keypoints for object deformation in a feed-forward manner. 
It is trained and evaluated using under the same setting as our method.
To incorporate more diverse representations, we also implemented an implicit rigging baseline based on  \textbf{NeuralDeformationGraph}~\cite{BozicPZTDN21}, a method that optimizes a neural deformation graph and per-node SDF field for dynamic shape reconstruction.
To adapt it to object rigging, we removed its time-dependent implicit shape prediction module.
In the first stage of its training, the deformation graph is optimized using frames from the training set.
In the second stage, the first frame of the test set is used to optimize the implicit shapes while keeping the graph fixed.
After training, we fix the implicit shape and optimize the graph's node positions and rotations to simulate pose editing.
We refer the readers to the supplementary materials for additional details.

\paragraph{Metrics.}
We use three metrics to evaluate the similarity between the target shape and the re-posed source shape: IoU for mesh similarity, and Chamfer L1 and L2 distances for point cloud similarity.
Since SkeRig and KeypointDeformer do not estimate surface meshes from input point clouds, we estimate the mesh surfaces for the \textit{refrigerator} and \textit{eyeglasses} datasets --- both of which lack ground-truth meshes --- using ground-truth occupancy values and Marching Cubes~\cite{LorensenC87} to compute IoU.

\paragraph{Results.}
We report quantitative comparison results for the five longest sequences from each of the five datasets in \cref{tab:comparison}.
The results show that our method outperforms all baselines by a large margin on nearly all datasets.
Qualitative comparisons in \cref{figure:comparison} demonstrates that our method accurately models object motion for both rigid and nonrigid dynamic object categories, and generates high-quality surfaces across a variety of shapes.
NeuralDeformationGraph\cite{BozicPZTDN21} produces noisy output, particularly for non-rigid objects, as it tends to overfit training shapes and lacks priors for modeling non-rigid deformations. In contrast, our method learns such priors in a data-driven manner, resulting in natural and accurate mesh surfaces after pose editing.
KeypointDeformer~\cite{Jakab0M0SK21} fails to capture intricate motions involving topological changes, as it deformes source shapes using cages. Our method mitigates this limitation by using a neural-network-based shape decoder.
SkeRig~\cite{Deng14} struggles to consistently optimize skeletons across different categories, often producing either over-simplified or overly complex skeletons that fail to capture object motion structure accurately.
On the contrary, our method automatically discovers blobs across all categories and consistently represents object motion with high accuracy.

\subsection{Animating the ``Clay-Monster''}
We have demonstrated the effectiveness and accuracy of our method in modeling object deformation on synthetic datasets.
To further showcase its potentials in real-world applications---particularly for casual users without 3D modeling expertise---we construct a small object category called \textit{"clay-monster"}, consisting of 12 clay figures scanned in 3 to 5 poses each using only an iPhone.
The scans are captured using ARKit 6~\cite{arkit}, and each takes approximately one minute to complete. 
Using this simple scanning pipeline, we are able to train the proposed model for pose manipulation of such artificial \textit{clay-monsters}, as illustrated in \cref{figure:rebuttal_clay}. This demonstrates how animatable representations can be easily obtained for object categories that lack standard rigging information using the proposed pipeline.
\input{fig/clay}

\subsection{Ablation Study}
\input{tab/ablation}
To validate the design choices of our method, we ablate important components from the proposed pipeline and analyze their impact on performance using the DeformingThings4D~\cite{LiTTZN21} dataset.
In particular, we ablate the identity conditioning operation and test the use of isotropic blobs to assess their contributions to output quality.
We also examine the effect of varying the numbers of blobs, $n_b$, where the original setting for DeformingThings4D uses $n_b=24$.
As shown in \cref{tab:ablation}, removing identity conditioning and anisotropic blobs significantly degrades model performance.
The results also indicate that increasing the number of blobs improves expressiveness of the model, albeit at the cost of higher computational cost.

%% file: fig/clay.tex
\begin{figure}[t]
    \centering
    \includegraphics[width=0.45\textwidth]{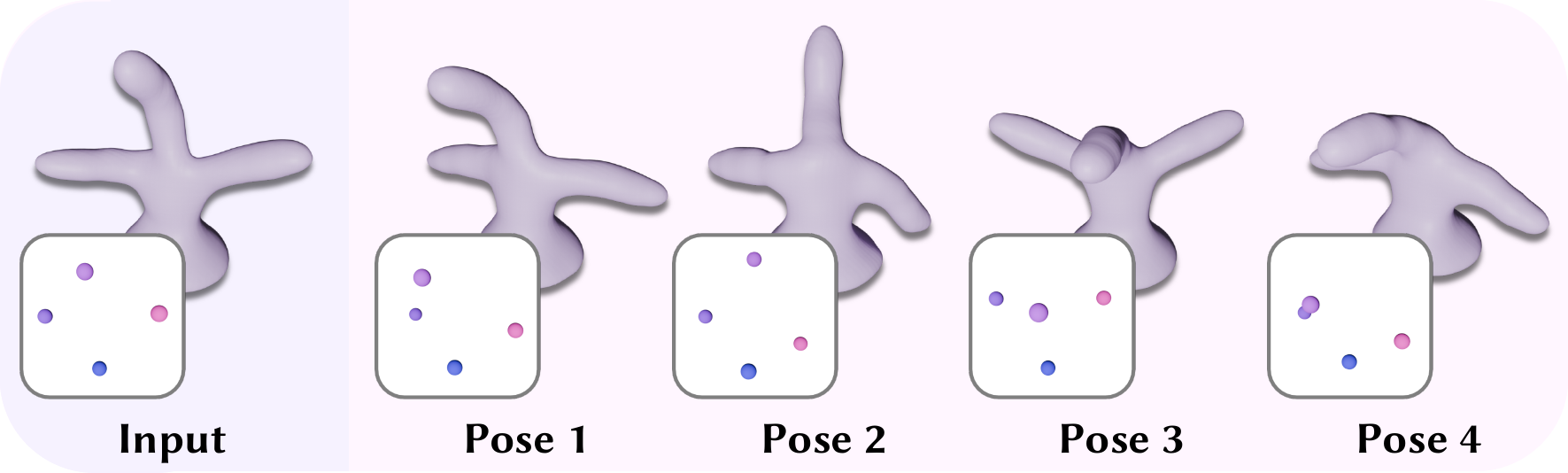}
    \caption{\textbf{Pose manipulation results} for a novel category (``clay-monster'') using our method, where no rigging tools are available.}
    \label{figure:rebuttal_clay}
\end{figure}

%% file: tab/ablation.tex
\begin{table}[t]
    \begin{center}
        
    \caption{%
        \textbf{Ablation Study.} We show ablation study on two different modules in our method as well as the impact of the number of blobs. All variants are trained on DeformingThings4D~\cite{LiTTZN21} dataset.
        Metrics are averaged over all sequences. IoU, Chamfer Distance $L_1$ and $L_2$ are reported.
    }
    \label{tab:ablation}
    \vspace{-7pt}
    \small
    \setlength{\tabcolsep}{6.5pt}
    \begin{tabular}{lccc} %
        \toprule
                                                   
                                                   & IoU $\uparrow$ & $CD_{1}$ $\downarrow$ & $CD_{2}$ $\downarrow$ \\
        \midrule
        \textit{w/o} Identity Conditioning       & 0.853              & 0.025             & 0.018     \\
        \textit{w/o} Anisotropic Blobs  & 0.934             & \cellfirst 0.017       & 0.012      \\
        \midrule
        Ours $K=8$ & 0.845 & 0.028 & 0.020 \\
        Ours $K=16$ & 0.927 & 0.020 & 0.013 \\
        \midrule
        Ours $K=24$                                      & \cellfirst 0.937   & \cellfirst 0.017    & \cellfirst 0.011       \\
        
        \bottomrule
    \end{tabular}
    \vspace{-20pt}
    \end{center}

\end{table}

%% file: sec/5_conclusions.tex
\section{Conclusion}
In this paper, we explored the novel task of learning category-specific rigging representations for dynamic objects in a category-agnostic manner. We introduced a data-driven pipeline for generic object rigging that can be readily applied to any object category.
Our method learns to disentangle object pose from identity by representing objects as a set of feature-embedded blobs in a fully unsupervised setting, and reconstruct surface meshes with rich geometric details from these blobs.
Experiments across five diverse datasets of distinct object categories demonstrate the effectiveness of our approach.